\documentclass{article}

\usepackage{arxiv}

\usepackage[utf8]{inputenc} 
\usepackage[T1]{fontenc}    
\usepackage{hyperref}       
\usepackage{url}            
\usepackage{booktabs}       
\usepackage{amsfonts}       
\usepackage{nicefrac}       
\usepackage{microtype}      
\usepackage{cleveref}       
\usepackage{lipsum}         
\usepackage{graphicx}
\usepackage{natbib}
\usepackage{doi}
\usepackage{multirow}

\title{Dual-Manifold Geometry Guided Representation Learning: Adaptive Coupling between Kernel and Data Spaces}

\newif\ifuniqueAffiliation
\uniqueAffiliationfalse

\ifuniqueAffiliation 
\author{ \href{https://orcid.org/0009-0005-9871-0960}{\includegraphics[scale=0.06]{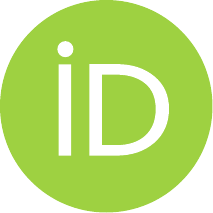}\hspace{1mm}Wencong Zhang}\thanks{These authors contributed equally to this work.} \\
	School of Biomedical Engineering\\
	Southern Medical University\\
	China, Guangzhou 510515 \\
	\texttt{zhangwc9926@gmail.com} \\
	\And
	\href{https://orcid.org/0009-0009-9051-8631}{\includegraphics[scale=0.06]{orcid.pdf}\hspace{1mm}Yue Zhang}\thanks{These authors contributed equally to this work.} \\
	School of Biomedical Engineering\\
	Southern Medical University\\
	China, Guangzhou 510515 \\
	\texttt{ashleyzhangsmu@gmail.com} \\
	\And
	\href{https://orcid.org/0000-0003-2138-9580}{\includegraphics[scale=0.06]{orcid.pdf}\hspace{1mm}Meiyan Huang} \\
	School of Biomedical Engineering\\
	Southern Medical University\\
	China, Guangzhou 510515 \\
	\texttt{huangmeiyan88@smu.edu.cn} \\
	\And
	\href{https://orcid.org/0000-0002-2161-3231}{\includegraphics[scale=0.06]{orcid.pdf}\hspace{1mm}Wei Yang} \\
	School of Biomedical Engineering\\
	Southern Medical University\\
	China, Guangzhou 510515 \\
	\texttt{weiyanggm@gmail.com} \\
	\And
	\href{https://orcid.org/0009-0008-8981-2238}{\includegraphics[scale=0.06]{orcid.pdf}\hspace{1mm}Qianjin Feng}\thanks{Corresponding author. E-mail: \texttt{fengqj99@smu.edu.cn}} \\
	School of Biomedical Engineering\\
	Southern Medical University\\
	China, Guangzhou 510515 \\
	\texttt{fengqj99@smu.edu.cn} \\
}
\else
\usepackage{authblk}

\newbox{\orcid}
\sbox{\orcid}{\includegraphics[scale=0.06]{orcid.pdf}} 

\author[1,2]{%
	\href{https://orcid.org/0009-0005-9871-0960}{\usebox{\orcid}\hspace{1mm}Wencong Zhang\textsuperscript{\dag}}%
}

\author[1,2]{%
	\href{https://orcid.org/0009-0009-9051-8631}{\usebox{\orcid}\hspace{1mm}Yue Zhang\textsuperscript{\dag}}%
}

\author[1,2]{%
	\href{https://orcid.org/0000-0003-2138-9580}{\usebox{\orcid}\hspace{1mm}Meiyan Huang}%
}

\author[1,2]{%
	\href{https://orcid.org/0000-0002-2161-3231}{\usebox{\orcid}\hspace{1mm}Wei Yang}%
}

\author[1,2]{%
	\href{https://orcid.org/0009-0008-8981-2238}{\usebox{\orcid}\hspace{1mm}Qianjin Feng\textsuperscript{*}}%
}
\affil[1]{School of Biomedical Engineering, Southern Medical University, China, Guangzhou 510515}
\affil[2]{Guangdong Provincial Key Laboratory of Medical Image Processing, Southern Medical University}

\fi

\renewcommand{\shorttitle}{\textit{arXiv} Template}

\hypersetup{
pdftitle={Dual-Manifold Geometry Guided Representation Learning:Adaptive Coupling between Kernel and Data Spaces},
pdfsubject={q-bio.NC, q-bio.QM},
pdfauthor={Qianjin Feng},
pdfkeywords={Dual-Manifold, Kernel-Guided, Exploit and Explore, Learnable Guidance Strength},
}

\begin{document}
\maketitle

\begingroup
\renewcommand{\thefootnote}{\fnsymbol{footnote}}  
\footnotetext[2]{These authors contributed equally to this work. E-mail: \texttt{zhangwc9926@gmail.com;ashleyzhangsmu@gmail.com}.}  
\footnotetext[1]{Corresponding author. E-mail: \texttt{fengqj99@smu.edu.cn}.} 
\endgroup

\begin{abstract}
	Deep representation learning has primarily focused on how feature representations evolve across network layers, while largely overlooking the structured geometry embedded in network parameters. In this work, we introduce a dual-manifold perspective of deep representation learning, in which each convolutional layer contains two coupled geometric spaces: a Kernel Manifold induced by convolutional filters and a Data Manifold characterized by intermediate feature representations. Since these two manifolds share the same channel space, the geometry of network parameters can provide complementary structural information for guiding feature evolution. Based on this insight, we propose Kernel-Guided Feature Transform (KGFT), a lightweight and nearly parameter-free module that derives a geometric guidance matrix from the kernel Gram matrix and uses it to transform the covariance structure of feature representations. Rather than reweighting feature responses as in conventional attention mechanisms, KGFT explicitly reshapes feature relationships by transferring geometric information from the kernel manifold to the data manifold. To accommodate the hierarchical nature of deep networks, we further introduce Exploit and Explore modes with a depth-aware scheduling strategy, together with a learnable guidance strength that adaptively controls the contribution of geometric transformation. This design promotes geometric alignment in shallow layers while encouraging feature diversity in deeper layers, without imposing excessive geometric constraints on representation learning. Theoretical analysis establishes the geometric validity of the proposed transformation and characterizes its effect on feature covariance. Extensive experiments across CNN- and Transformer-based architectures, including ResNet, ViT, and LLaMA-7B, demonstrate consistent improvements on image classification and arithmetic reasoning tasks, validating the generality and effectiveness of kernel-guided dual-manifold representation learning. Code is available at \url{https://github.com/ZWC-SMU/KGFT}.
\end{abstract}


\section{Introduction}
Deep neural networks (DNN) have achieved remarkable success across a wide range of text and visual tasks by progressively transforming low-level patterns into high-level semantic representations \citep{1krizhevsky2012imagenet,2simonyan2014very,3szegedy2015going,4he2016deep,5howard2017mobilenets,6tan2019efficientnet,7vaswani2017attention,8dosovitskiy2020image,9gu2023mamba,10liu2024vmamba}. A fundamental principle underlying this success is hierarchical representation learning, where shallow layers capture basic structures such as edges and textures, while deeper layers encode increasingly abstract semantic concepts. Consequently, most existing studies focus on understanding and improving how feature representations evolve across network layers through architectural designs \citep{4he2016deep,7vaswani2017attention,9gu2023mamba}, representation learning \citep{11hinton1993autoencoders,12mikolov2013efficient,13chen2020simple,14he2022masked}, and attention mechanisms \citep{7vaswani2017attention,15hu2018squeeze,16woo2018cbam,17liu2021swin}. Although the representations extracted by these methods have achieved impressive performance in downstream tasks such as classification, detection, segmentation, and generation, they predominantly characterize representation learning from the feature space, treating intermediate features as the primary carriers of information and network parameters as transformation operators. However, features are not learned in isolation, they are generated through the interaction between input data and the parameters that transform them. In particular, convolutional kernels undergo continuous adaptation during training and gradually develop non-random correlations and structured configurations. Recent studies have revealed that network weights exhibit geometric and spectral structures, which are closely associated with the evolution and organization of learned features \citep{18beaglehole2024feature,19defilippis2026scaling,20cha2026weight}. These findings suggest that the parameter space itself contains meaningful structural information that complements the geometry of the feature space.

This observation motivates us to reconsider representation learning from a joint geometric perspective: \textbf{Can the geometry encoded by network parameters actively guide the evolution of feature representations?} Within a convolutional layer, the representation process can be viewed through two coupled geometric structures. On the one hand, convolutional weights induce a kernel geometry, where correlations among filters characterize the relationships encoded by the learned parameters. On the other hand, feature representations induce a data geometry, whose covariance structure describes how feature channels are organized in the representation space \citep{21raghu2017svcca,22kornblith2019similarity}. Although generated from different sources, the two manifolds share the same channel space and co-evolve throughout network optimization. This intrinsic correspondence suggests that the kernel manifold provides an informative geometric prior for guiding the evolution of the data manifold. However, existing representation learning methods typically optimize feature representations while overlooking this dual-manifold interaction, leaving the relationship between parameter geometry and feature geometry largely unexplored.

Motivated by above, we propose a novel \textbf{Dual-Manifold Geometry Guided Representation Learning} framework, which formulates deep representation learning as a dynamic coupling process between kernel geometry and data geometry. Instead of reweighting feature responses as in conventional attention mechanisms, our method transfers structural information from the kernel manifold to reshape the covariance structure of feature representations. This geometric interaction is implemented by a lightweight \textbf{Kernel-Guided Feature Transform (KGFT)} module, which derives a kernel-induced guidance matrix and employs a learnable guidance strength to adaptively control the contribution of geometric transformation. To accommodate the hierarchical learning behavior of deep networks, we further introduce a dual-mode guidance strategy, namely \textbf{Exploit and Explore}. Specifically, shallow layers emphasize geometric alignment to establish stable low-level representations, whereas deeper layers encourage geometric expansion to promote semantic diversity. This depth-aware scheduling enables kernel geometry to guide representation learning throughout the entire optimization process. Extensive experiments across both CNN- and Transformer-based architectures, including ResNet, ViT, and LLaMA-7B, demonstrate that KGFT consistently improves performance on image classification and arithmetic reasoning tasks while introducing negligible additional overhead. The proposed framework is simple, lightweight, and can be readily incorporated into existing convolutional networks without modifying their backbone architectures. Our contributions can be summarized as the followings:
\begin{itemize}
	\item A dual-manifold representation learning framework. We formulate feature learning as the geometric interaction between a kernel manifold and a data manifold, providing a new perspective on representation learning in deep neural networks.
	\item A lightweight kernel-guided feature transformation. We propose KGFT as a plug-and-play module with negligible parameter overhead that transfers geometric structures from convolutional kernels to feature representations through covariance transformation, enabling adaptive geometry-guided feature learning.
	\item A depth-aware dual-mode scheduling strategy. We design a simple yet effective geometry scheduling mechanism that performs geometric alignment in shallow layers (Exploit) and geometric expansion in deep layers (Explore).
	\item Extensive experimental validation. We demonstrate its effectiveness across CNN- and Transformer-based architectures, including multiple ResNet variants, ViT-Tiny, and LLaMA-7B, on image classification and arithmetic reasoning tasks.
\end{itemize}

\section{Related work}
\subsection{Manifold Perspective of Deep Neural Networks}
Deep neural networks can be viewed as a sequence of nonlinear transformations that progressively reshape the underlying data manifold \citep{21raghu2017svcca,23bengio2013representation}. A growing body of work has investigated this geometric evolution from different perspectives, including representation dynamics, manifold geometry, and Riemannian formulations \citep{24cohen2020separability,25hauser2017principles,26benfenati2023singular}. Despite these advances, existing manifold-based analyses mainly focus on the evolution of data manifolds induced by feature transformations, while the intrinsic geometric structures embedded in network parameters remain less explored. Recent studies have revealed that network parameters also induce structured geometric representations through kernel formulations. Jacot et al. \citep{27jacot2018neural} introduced the Neural Tangent Kernel (NTK), demonstrating that the parameter space of neural networks implicitly defines a kernel geometry that governs the evolution of learned functions. Furthermore, Fort et al. \citep{28fort2020deep} showed that the induced kernel is not static during training but continuously evolves with parameter updates, highlighting the dynamic geometric relationship between network parameters and learned representations. However, existing studies primarily investigate parameter-induced kernel geometry and feature-space geometry independently. In contrast, our work views a deep convolutional network as a coupled system of two interacting manifolds: the kernel manifold defined by convolutional weights and the data manifold represented by feature distributions. By explicitly modeling their geometric relationship, we investigate how parameter geometry can guide feature evolution.

\subsection{Feature Covariance Modeling}
Feature covariance provides a fundamental means of characterizing second-order relationships among feature representations. Such second-order information has been shown to contain complementary and often highly discriminative information beyond individual feature responses \citep{29gatys2016image,30kong2017low,31li2017second,32wang2026towards}. For example, Gram matrix based methods, such as neural style transfer \citep{29gatys2016image}, demonstrated that covariance statistics capture meaningful feature structures. Subsequent covariance pooling approaches \citep{33li2018towards} further exploited second-order statistics for compact representation learning. However, existing covariance-based methods mainly treat feature statistics as the target representation to be normalized or aggregated. In contrast, our method introduces a different perspective: the covariance structure of features is not only analyzed but actively modulated by the geometric structure of convolutional kernels.

\subsection{Kernel Methods and Manifold Mapping}
Kernel methods provide a principled way to characterize geometric relationships through kernel functions and their associated Gram matrices \citep{20cha2026weight,scholkopf1998nonlinear,esser2024non}. Recent studies have extended this geometric perspective to deep neural networks, showing that learned representations exhibit structured manifold geometry and evolve systematically across network layers \citep{scholkopf1998nonlinear,esser2024non}. In particular, the weight Gram matrix has been shown to capture meaningful structures underlying feature dynamics, revealing a close relationship between parameter geometry and the evolution of learned representations \citep{20cha2026weight}. These findings suggest that kernel-induced geometry can provide informative structural priors for representation learning. Inspired by these principles, our method interprets both convolutional kernels and feature activations as geometric objects. 

\subsection{Attention Mechanisms and Feature Recalibration}
Attention mechanisms have become a fundamental paradigm for improving feature representation by dynamically modeling the relative importance or relationships among different elements \citep{15hu2018squeeze,16woo2018cbam,li2019selective,37wang2020eca}. Channel- and spatial-attention methods, such as SENet \citep{15hu2018squeeze}, CBAM \citep{16woo2018cbam}, SKNet \citep{li2019selective}, and ECA-Net \citep{37wang2020eca}, adapt feature responses through channel recalibration or selective aggregation, while self-attention extends this principle to pairwise interactions among tokens \citep{7vaswani2017attention,8dosovitskiy2020image,17liu2021swin}. These methods demonstrate that dynamically modeling feature relationships can effectively enhance representation quality. However, existing attention mechanisms primarily operate within the feature space, emphasizing or aggregating features according to their learned responses. In contrast, KGFT introduces kernel-guided geometric adaptation: rather than reweighting feature elements, it transfers the structural geometry of convolutional kernels to reshape feature relationships.  

\section{Method}
\subsection{Dual-Manifold Formulation}
Deep neural networks can be interpreted as a hierarchical geometric transformation process, where feature distributions evolve across network depth. Existing manifold-based analyses mainly focus on the geometry of learned features, while ignoring the intrinsic geometric structures encoded in network parameters. To bridge these two perspectives, we propose a dual-manifold formulation, which models a convolutional layer as an interaction between two coupled manifolds: the kernel-manifold induced by convolutional weights and the data-manifold formed by feature representations.

\subsubsection{Kernel Manifold}
Let $\mathbf{W}\in\mathbb{R}^{C\times D}$ be the weight matrix of a certain convolutional layer, where $C$ denotes the number of output channels and $D=k^2C_{in}$ represents the flattened kernel dimension. Treating each row of $\mathbf{W}$ as a point in $\mathbb{R^D}$, these $C$ points span a kernel-manifold $\mathbf{\mathcal{M}}_{\mathbf{W}}\subset\mathbb{R}^{D}$. We characterize the intrinsic geometric relationships through the kernel Gram matrix:
\begin{equation}
	\mathbf{G} = \mathbf{W}\mathbf{W}^\top \in\mathbb{R}^{C\times C} 
\end{equation}
The element $\mathbf{G}_{ij}=\langle\mathbf{w}_{i},\mathbf{w}_{j}\rangle$ is the inner-product similarity between the $i$-th and $j$-th kernels, reflecting their angular relationship and correlation structure. The eigen-decomposition of $\mathbf{G}=\mathbf{WW}^\top\in\mathbb{R}^{C\times C}$ reveals the principal geometric directions of the kernel manifold. Larger eigenvalues indicate that the network allocates stronger representational capacity along the corresponding kernel directions.

\subsubsection{Data Manifold}
Let the input features for the current layer be $\mathbf{X}\in\mathbb{R}^{B\times C\times H \times W}$, flattened into $\mathbf{X_f}\in\mathbb{R}^{N\times C}$ (where $N=B\times H \times W$). Each spatial feature vector is considered as a point in the channel space, forming the data manifold: $\mathbf{\mathcal{M}}_{\mathbf{X}}\subset\mathbb{R}^{C}$. The geometric structure of  is characterized by the feature covariance matrix:
\begin{equation}
	\mathbf{K}=\frac{1}{\sqrt{N}}\mathbf{X_{f}}^{\top}\mathbf{X_{f}}\in\mathbb{R}^{C\times C}
\end{equation}
The element $\mathbf{K}_{ij}=\langle\mathbf{k}_{i},\mathbf{k}_{j}\rangle$ represents the covariance relationship between the $i$-th and $j$-th feature channels, describing the intrinsic distribution of learned representations.

\subsubsection{Dual-Manifold Correspondence}
The kernel manifold $\mathbf{\mathcal{M}}_{\mathbf{W}}$ is constructed from convolutional filters, whereas the data manifold $\mathbf{\mathcal{M}}_{\mathbf{X}}$ is formed by spatial feature vectors. Although they originate from different spaces, their geometric structures are represented in the same channel space. Accordingly, the kernel Gram matrix $\mathbf{G}$ characterizes how filters are organized and correlated in parameter space, while the feature covariance matrix $\mathbf{K}$ describes how feature responses are distributed and correlated in data space. Thus, the kernel geometry provides structural cues about dominant and weak channel directions, which can be transferred to modulate the corresponding feature geometry. To establish this interaction, we introduce a kernel-induced geometric guidance matrix $\mathbf{M}$, which adaptively enhances or suppresses feature directions according to the underlying kernel geometry. The construction of $\mathbf{M}$ is described in the following section.

\subsection{Dual-Manifold Geometric Guidance}
The goal of KGFT is to transfer kernel geometry into feature space. However, different network depths require different geometric behaviors. Shallow layers require stable feature construction, while deep layers require semantic diversity. Therefore, we design two complementary guidance modes:
\begin{itemize}
\item Exploit mode: preserve and reinforce reliable kernel structures;
\item Explore mode: suppress excessive correlations and encourage new feature directions.
\end{itemize}

\subsubsection{Dual-Mode Guidance Matrix}
We first get the guidance matrix $\mathbf{M}\in\mathbb{R}^{C\times C}$ from the metric $\mathbf{G}$ of the kernel manifold, whose $\mathbf{M}$ includes Exploit and Explore modes:

\textbf{Exploit Mode:} For shallow layers, stable low-level representations are more important than representation diversity. Therefore, we directly utilize the kernel geometry:
\begin{equation}
	\mathbf{M}_\mathrm{exploit}=\mathbf{G}+\varepsilon \mathbf{I_C}
\end{equation}
where $\varepsilon > 0$ ensures positive definiteness. Consequently, directions strongly represented by the kernel manifold receive larger geometric emphasis, whereas weak kernel directions receive relatively smaller emphasis. The geometric interpretation and positive definiteness of $\mathbf{M}_\mathrm{exploit}$ is formally established in Appendix A. 6.1 and 6.2.
	
\textbf{Explore Mode:} As network depth increases, excessive dependence on existing kernel correlations may restrict semantic diversity. To encourage feature exploration, we first normalize the kernel Gram matrix:
\begin{equation}
	\mathbf{C=D^{-1/2}GD^{-1/2}}, \, where \, \mathbf{D}=diag(\mathbf{G})
\end{equation}
The explore guidance matrix is then defined as:
\begin{equation}
	\mathbf{M}_{\mathrm{explore}}=\delta(\mathbf{I}_C-\mathbf{C})+\varepsilon\mathbf{I}_C
\end{equation}
where $\delta=\frac{1}{C}\Sigma_{i=1}^C\mathbf{G}_{ii}$ denotes the average geometric scale. Different from exploit mode, explore mode suppresses highly correlated kernel directions while amplifying weakly correlated directions, encouraging feature representations to expand toward unexplored semantic dimensions. The geometric interpretation and positive definiteness of $\mathbf{M}_{\mathrm{explore}}$ is formally established in Appendix A. 6.1 and 6.2.

\subsubsection{Kernel-Guided Feature Transformation}
After obtaining the geometric guidance matrix: $\mathbf{M}\in{\mathbf{M}_\mathrm{exploit}, \mathbf{M}_{\mathrm{explore}}}$, KGFT uses $\mathbf{M}$ to modulate the channel geometry of the feature representation:
\begin{equation}
	\mathbf{K'}=\mathbf{MK}\in\mathbb{R}^{C\times C}
\end{equation}
The new metric $\mathbf{K'}$ is applied to the feature transformation:
\begin{equation}
	\mathbf{Y_{f}=X_{f}\mathbf{K'}=X_{f}M\mathbf{K}}
\end{equation}
which the transformed feature representation assigns different geometric weights to different channel directions according to the kernel structure. The transformed feature is then reshaped back to its original spatial dimensions and processed by a lightweight projection layer:
\begin{equation}
	\mathbf{Y}=LN(\mathrm{Proj}(\mathrm{reshape}(\mathbf{Y_f})))
\end{equation}
where $\mathrm{Proj}$ and $LN$ denote a $1\times 1$ convolution and layer normalization, respectively. Finally, the transformed feature is fused with the original representation through residual learning:
\begin{equation}
	\mathbf{Y}=\mathbf{X} + \mathbf{Y}
\end{equation}
The residual connection preserves the original feature pathway, while the transformed branch introduces adaptive geometric guidance from the kernel manifold. The stability of the mapping from kernel parameters to the resulting feature transformation is established in Appendix A. 6.3.

\subsubsection{Adaptive Geometry Guidance Strength}
Although kernel geometry provides meaningful structural priors, excessive guidance may constrain early representation learning. Therefore, KGFT introduces a learnable geometry guidance strength allows the network to automatically determine the contribution of geometric transformation. Specifically, we parameterize the guidance strength as a trainable scalar:
\begin{equation}
	\mathbf{s}=\sigma(\alpha)
\end{equation}
where $\alpha$ is a learnable parameter and $\sigma(\cdot)$ denotes the sigmoid function that maps the guidance strength to the interval [0,1]. During training, a warm-up strategy is adopted to gradually activate the learnable geometric guidance, thereby preventing excessive geometric perturbation in the early optimization stage. The final KGFT transformation is formulated as:
\begin{equation}
	\mathbf{Y}=\mathbf{X}+\mathbf{s}\cdot\mathcal{F}_{KGFT}(\mathbf{X})
\end{equation}
where $\mathcal{F}_{KGFT}(\cdot)$ denotes the geometry-guided feature transformation described in Section 3.2.2. The learnable coefficient $\mathbf{s}$ adaptively controls the extent to which the transformed feature representation contributes to the output. When $\mathbf{s}\to0$, KGFT approaches an identity mapping, allowing the network to preserve the original feature representation. As $\mathbf{s}\to1$, the geometric transformation is fully incorporated, enabling the network to maximally exploit the selected geometric guidance. This bounded residual formulation provides an explicit mechanism for controlling the magnitude of geometric modulation. A formal boundedness and identity-preserving analysis is provided in Appendix A. 6.4.

\subsection{Dual-Manifold Scheduling in Deep Neural Networks}
To apply KGFT to deep neural architectures, we introduce a depth-aware dual-manifold scheduling strategy that determines where and how KGFT modules are incorporated into the network. The underlying principle is that shallow layers perform geometry alignment (Exploit), whereas deeper layers perform geometry exploration (Explore). We apply this scheduling strategy to both convolutional and Transformer-based architectures, including ResNet, ViT, and LLaMA-7B. For ResNet architectures, KGFT modules are inserted at selected stages following the shallow-to-deep transition. For ViT and LLaMA-7B, which consists of a sequence of Transformer encoder layers, we similarly adopt a sparse layer-wise scheduling strategy. The detailed configurations for different ResNet and Transformer architectures are summarized in the Table \ref{table5} and Fig. \ref{fig2} of Appendix B.

\section{Experiments}
\subsection{Datasets and Metrics}
We evaluate KGFT across both vision and language tasks to assess its generality beyond convolutional architectures. For image classification, we use two widely adopted benchmarks, CIFAR-100 \citep{krizhevsky2009learning}
and ImageNet-1K \citep{39deng2009imagenet}. CIFAR-100 contains 50,000 training and 10,000 test images from 100 classes, with a spatial resolution of $32\times 32$. ImageNet-1K contains approximately 1.28 million training images and 50,000 validation images across 1,000 classes. Images are randomly resized and cropped to $224\times 224$ during training. Following standard practice, we apply random cropping, horizontal flipping, label smoothing, and Cutout augmentation \citep{40devries2017improved}. Top-1 classification accuracy is used as the evaluation metric. Furthermore, we conduct arithmetic reasoning experiments with LLaMA-7B \citep{41touvron2023llama}, using MATH10K \citep{42wu2024reft} for training and evaluating the fine-tuned models on four held-out benchmarks, including AQuA \citep{43ling2017program}, GSM8K \citep{44cobbe2021training}, MAWPS \citep{45koncel2016mawps}, and SVAMP \citep{46patel2021nlp}. For these language tasks, answer accuracy is used as the evaluation metric.

\subsection{Experimental Settings}
We evaluate the proposed KGFT on representative residual architectures with different depths and model capacities, including ResNet20 (0.27M parameters, 3 stages), ResNet32 (0.47M parameters, 3 stages), ResNet18 (11.15M parameters, 4 stages), ResNet34 (20.79M parameters, 4 stages), and ResNet50 (24.37M parameters, 4 stages). These models cover both lightweight networks and deeper architectures, enabling a comprehensive evaluation of the effectiveness and scalability of KGFT. 

To adapt KGFT to Transformer architectures, we generalize the channel-wise correspondence used in convolutional networks to the hidden representation space of Transformer blocks. Specifically, for the $l$-th Transformer block, let $\mathbf{W}^{down}_{l}\in\mathbb{R}^{H\times I}$ denote the weight matrix of the MLP down-projection layer, where $H$ and $I$ denote the hidden dimension and the intermediate MLP dimension, respectively. The down-projection maps the intermediate MLP representation back to the $H$-dimensional hidden space, each row of $\mathbf{W}^{down}_{l}$ corresponds to one output hidden dimension. We therefore construct the hidden-space kernel geometry as
\begin{equation}
	\mathbf{G}_1 = \mathbf{W}_1^{down} \left(\mathbf{W}_1^{down} \right)^T \in \mathbb{R}^{H \times H}
\end{equation}
Similarly, in LoRA \citep{hu2021lora} fine-tuning LLaMA-7B, we define the Transformer Kernel Manifold as
\begin{equation}
	\mathbf{G}_1 = \mathbf{W}_1^{eff} \left(\mathbf{W}_1^{eff} \right)^T \in \mathbb{R}^{H \times H}
\end{equation}
where $\mathbf{W}_1^{eff}$ denotes the effective down-projection weight during LoRA fine-tuning. Notably, for both ViT and LLaMA-7B, the final projection layer in KGFT is removed when adapting the module to Transformer architectures.

For optimization, CIFAR-100 models are trained for 200 epochs with a batch size of 512, while ImageNet-1K models are trained for 100 epochs with a batch size of 256. All experiments use stochastic gradient descent (SGD) with a momentum of 0.9 and an initial learning rate of 0.1. The learning rate is scheduled by MultiStepLR, with decay factors of 0.1 applied at 30\%, 60\%, and 80\% of the total training epochs. The weight decay is set to $5 \times 10^{-4}$, and label smoothing with a coefficient of 0.01 is adopted for regularization. For the arithmetic reasoning experiments, we fine-tune LLaMA-7B on the Math10K training set using LoRA for 3 epochs. We use the AdamW optimizer with an initial learning rate of $3 \times 10^{-4}$ and a batch size of 16. The learning rate follows a linear decay schedule with 100 warm-up steps.

For KGFT, we employ the proposed dual-mode scheduling strategy, where Exploit guidance is applied to shallow stages to preserve discriminative local structures, while Explore guidance is introduced in the final stage to enhance global feature exploration. The learnable-strength parameter is optimized with a smaller learning rate ($0.5 \times$ the base learning rate) and a 50-epoch warm-up strategy.

To ensure reliable evaluation, all CIFAR-100 experiments are repeated using eight different random seeds, ImageNet-1K experiments are conducted with four independent runs, and MATH10K experiments are employed with four independent runs. The reported results are averaged across multiple runs.

\subsection{Results}
\subsubsection{Image Classification on CIFAR-100}
Table \ref{table1} presents the Top-1 accuracy results of KGFT on CIFAR-100 across different ResNet architectures. All results are obtained using eight different random seeds, with the mean accuracy, standard deviation, best accuracy, and worst accuracy reported. Compared with the standard baselines, KGFT consistently improves classification performance with negligible parameter overhead. Specifically, KGFT improves the mean accuracy by 1.78, 1.16, and 2.24 percentage points on ResNet20, ResNet32, and ResNet18, respectively. Notably, on ResNet18, KGFT improves the mean accuracy from 74.60 to 76.84 while reducing the standard deviation from 0.79 to 0.26, indicating substantially improved optimization stability across different random initializations. These consistent improvements demonstrate the effectiveness of adaptively incorporating geometric guidance into representation learning. The results further suggest that the geometric information captured by the kernel and data manifolds provides beneficial structural guidance for feature learning, while the learnable guidance strength allows the network to adaptively regulate the contribution of geometric transformation during optimization.

\begin{table}[t]
	\centering
	\caption{Top-1 classification accuracy (\%) on CIFAR-100 across eight different random seeds for different ResNet architectures and KGFT variants. Mean, standard deviation, best, and worst accuracies are reported.}
	\label{table1}
	\renewcommand{\arraystretch}{1.5}  
	\begin{tabular}{l l r c c c c}
		\toprule
		\textbf{Model} & \textbf{Methods} & \textbf{Params} & \textbf{Mean} & \textbf{Std} & \textbf{Best} & \textbf{Worst} \\
		\midrule
		\multirow{2}{*}{ResNet-20} & Baseline & 0.27M & 67.61 & 0.28 & 68.11 & 67.21 \\
		& + KGFT   & 0.28M & \textbf{69.39} & \textbf{0.16} & \textbf{69.60} & \textbf{69.18} \\
		\midrule
		\multirow{2}{*}{ResNet-32} & Baseline & 0.47M & 69.69 & \textbf{0.36} & 70.09 & 68.98 \\
		& + KGFT   & 0.48M & \textbf{70.85} & 0.55 & \textbf{71.80} & \textbf{69.76} \\
		\midrule
		\multirow{2}{*}{ResNet-18} & Baseline & 11.15M & 74.60 & 0.79 & 75.38 & 72.78 \\
		& + KGFT   & 11.48M & \textbf{76.84} & \textbf{0.26} & \textbf{77.25} & \textbf{76.50} \\
		\bottomrule
	\end{tabular}
\end{table}

\begin{table}[t]
	\centering
	\caption{Top-1 classification accuracy (\%) on CIFAR-100 across eight different random seeds for ViT-tiny architectures and KGFT. Mean, standard deviation, best, and worst accuracies are reported.}
	\label{table2}
	\renewcommand{\arraystretch}{1.5}  
	\begin{tabular}{l l r c c c c}
		\toprule
		\textbf{Model} & \textbf{Methods} & \textbf{Params} & \textbf{Mean} & \textbf{Std} & \textbf{Best} & \textbf{Worst} \\
		\midrule
		ViT-tiny & Baseline & 5.54M & 54.33 & \textbf{0.18} & 54.42 & 54.11 \\
		ViT-tiny & + KGFT   & 5.54M & \textbf{54.96} & 0.48 & \textbf{55.20} & \textbf{54.58} \\
		\bottomrule
	\end{tabular}
\end{table}

We also apply KGFT to the ViT-tiny. As shown in Table \ref{table2}, KGFT improves the mean Top-1 accuracy from 54.33 to 54.96, yielding a 0.63 percentage-point gain with no additional trainable parameters. Although the improvement is more modest than that observed on the ResNet architectures, the consistent accuracy gain demonstrates that manifold-guided feature transformation can also benefit Transformer-based visual representations. Overall, the results support the central idea of KGFT: explicitly coupling kernel and data manifold geometry can provide effective complementary guidance for representation learning, and a learnable guidance strength enables the network to adaptively exploit this geometric information according to its evolving optimization state.

\begin{table}[t]
	\centering
	\caption{Top-1 classification accuracy (\%) on ImageNet-1K across four different random seeds for different ResNet architectures and learnable-strength KGFT. Mean, standard deviation, best, and worst accuracies are reported.}
	\label{table3}
	\renewcommand{\arraystretch}{1.5}  
	\begin{tabular}{l l r c c c c}
		\toprule
		\textbf{Model} & \textbf{Methods} & \textbf{Params} & \textbf{Mean} & \textbf{Std} & \textbf{Best} & \textbf{Worst} \\
		\midrule
		ResNet-34 & Baseline & 20.79M & 73.57 & \textbf{0.18} & 73.84 & 73.46 \\
		ResNet-34 & + KGFT   & 21.20M & \textbf{74.07} & 0.48 & \textbf{74.55} & \textbf{73.41} \\
		\midrule
		ResNet-50 & Baseline & 24.37M & 75.43 & 0.31 & 75.70 & 75.06 \\
		ResNet-50 & + KGFT   & 24.79M & \textbf{76.58} & \textbf{0.16} & \textbf{76.81} & \textbf{76.42} \\
		\bottomrule
	\end{tabular}
\end{table}

\begin{figure*}[t]
	\centering
	\includegraphics[width=0.95\columnwidth]{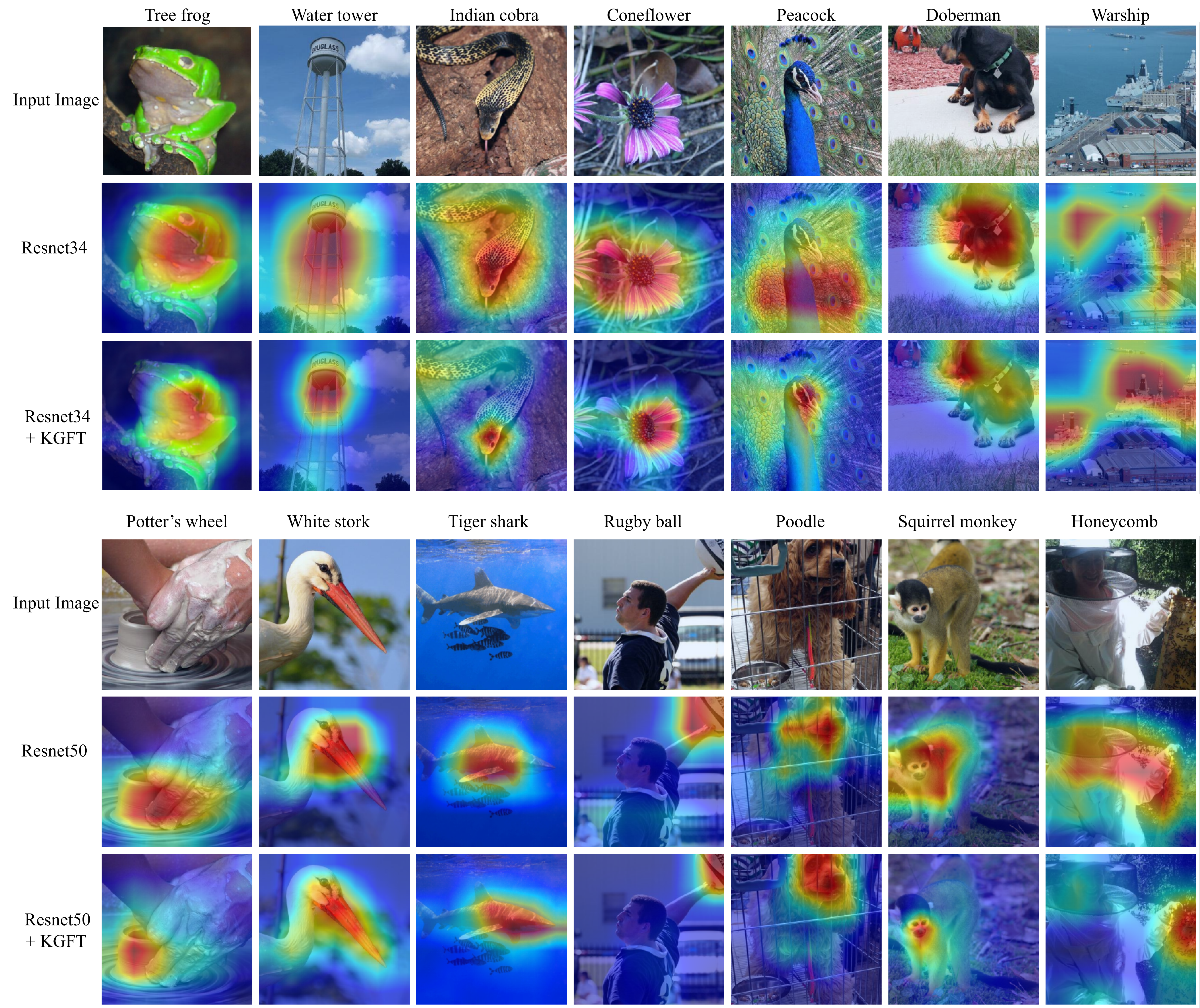}
	\caption{Grad-CAM \citep{49selvaraju2017grad} visualization of baseline ResNet models and their KGFT-enhanced counterparts on ImageNet-1K. KGFT generally produces more localized responses on class-relevant regions while suppressing irrelevant background activations.}
	\label{fig1}
\end{figure*}

\subsubsection{Image Classification on ImageNet-1K}
Table \ref{table3} further evaluates KGFT on the more challenging ImageNet-1K benchmark using ResNet-34 and ResNet-50. KGFT consistently improves the Top-1 accuracy over the corresponding baselines, with gains of 0.50 and 1.15 percentage points on ResNet-34 and ResNet-50, respectively. The larger improvement on ResNet-50 suggests that the benefit of geometric guidance becomes more evident as the representation space becomes more complex with increasing model capacity. KGFT also improves the robustness of training, particularly for ResNet-50, where the standard deviation decreases from 0.31 to 0.16 and the worst-case accuracy increases from 75.06 to 76.42. These results demonstrate that the effectiveness of KGFT is not limited to small-scale CIFAR-100 experiments but extends to large-scale ImageNet-1K classification. More importantly, the consistent gains across architectures support our underlying hypothesis that kernel- and data-manifold geometry provides complementary structural information beyond conventional feature learning, while adaptive coupling allows this information to be effectively integrated into networks with different representation capacities. Fig. \ref{fig1} also presents Grad-CAM activation maps of ResNet-34 and ResNet-50 with and without KGFT. Compared with the baselines, KGFT produces more localized and class-relevant responses, with reduced activation on irrelevant backgrounds. This observation suggests that kernel-guided geometric transformation promotes more discriminative feature representations.

\begin{table}
	\centering
	\caption{Accuracy (\%) on four benchmarks across four different random seeds for LLaMA-7B and learnable-strength KGFT. *Performance results of all baseline methods are taken from Wu et al \citep{41touvron2023llama}. Mean, standard deviation, best, and worst accuracies are reported.}
	\label{table4}
	\renewcommand{\arraystretch}{1.5}  
	\begin{tabular}{l l r l c c c c}
		\toprule
		\textbf{Model} & \textbf{Methods} & \textbf{Params (\%)} & \textbf{Testsets} & \textbf{Mean} & \textbf{Std} & \textbf{Best} & \textbf{Worst} \\
		\midrule
		LLaMA-7B & Baseline (LoRA) & 0.8256M & GSM8K & 37.50 & -- & -- & -- \\
		LLaMA-7B & Baseline (LoRA) & 0.8256M & MAWPS  & 79.00 & -- & -- & -- \\
		LLaMA-7B & Baseline (LoRA) & 0.8256M & SVAMP  & 52.10 & -- & -- & -- \\
		LLaMA-7B & Baseline (LoRA) & 0.8256M & AQuA   & \textbf{18.90} & -- & -- & -- \\
		\midrule
		LLaMA-7B & + KGFT (L9) & 0.8256M & GSM8K  & \textbf{38.32} & 0.64 & 39.42 & 37.83 \\
		LLaMA-7B & + KGFT (L9) & 0.8256M & MAWPS  & 82.46 & 0.35 & 82.77 & 81.93 \\
		LLaMA-7B & + KGFT (L9) & 0.8256M & SVAMP  & \textbf{53.83} & 1.76 & 56.30 & 51.60 \\
		LLaMA-7B & + KGFT (L9) & 0.8256M & AQuA   & 15.95 & 2.37 & 19.29 & 12.60 \\
		\midrule
		LLaMA-7B & + KGFT (L9/L21/L32) & 0.8256M & GSM8K  & 37.91 & 0.66 & 38.74 & 37.00 \\
		LLaMA-7B & + KGFT (L9/L21/L33) & 0.8256M & MAWPS  & \textbf{82.77} & 1.36 & 84.87 & 81.09 \\
		LLaMA-7B & + KGFT (L9/L21/L32) & 0.8256M & SVAMP  & 53.30 & 1.69 & 55.20 & 50.70 \\
		LLaMA-7B & + KGFT (L9/L21/L32) & 0.8256M & AQuA   & 16.93 & 2.21 & 19.69 & 14.57 \\
		\bottomrule
	\end{tabular}
\end{table}

\subsubsection{Arithmetic Reasoning with LLaMA-7B}
To investigate whether the proposed learnable strength KGFT can generalize beyond convolutional architectures, we extend KGFT to the Transformer-based LLaMA-7B model and evaluate its effectiveness on arithmetic reasoning tasks. We fine-tune LLaMA-7B on MATH10K, a combined training set constructed from seven arithmetic reasoning datasets with language-model-generated chain-of-thought rationales. We report results on four mathematical reasoning benchmarks: AQuA \citep{43ling2017program}, GSM8K \citep{44cobbe2021training}, MAWPS \citep{45koncel2016mawps}, and SVAMP \citep{46patel2021nlp}. Models need to generate chain-of-thought \citep{48wei2022chain} before the final answer. Following standard evaluation practice, only the correctness of the final numerical or multiple-choice answer is considered when computing accuracy.

We adopt LoRA as the parameter-efficient fine-tuning baseline and follow the training configuration of Wu et al. \citep{42wu2024reft}. LoRA represents task-specific weight updates using low-rank matrices while keeping the pretrained model parameters frozen. LLaMA-7B consists of 32 Transformer decoder blocks. We evaluate two KGFT insertion strategies. In the first setting, a single KGFT module in exploit mode is inserted into decoder block 9. In the second setting, KGFT modules are inserted into decoder blocks 9, 21, and 32, using exploit modes.

As shown in Table \ref{table4}, KGFT generalizes beyond CNN architectures to Transformer-based large language models. On GSM8K, MAWPS, and SVAMP, both single-layer and multi-layer KGFT outperform the LoRA baseline. Specifically, single-layer KGFT yields improvements of 0.82, 3.46, and 1.73, respectively, while multi-layer KGFT achieves gains of 0.41, 3.77, and 1.20. These results demonstrate that kernel-guided geometric adaptation can effectively guide hidden representations in language models, suggesting that the proposed dual-manifold interaction is not restricted to convolutional feature spaces. Nevertheless, the gains vary across reasoning benchmarks. In particular, KGFT achieves lower mean accuracy than LoRA on AQuA, indicating that the effectiveness of geometric guidance is task-dependent rather than uniformly beneficial. This variation may arise from differences in task-specific data distributions and reasoning characteristics, highlighting the need for adaptive geometric guidance across different reasoning tasks.

\section{Conclusion}
Deep representation learning has predominantly focused on the geometry of intermediate features, while overlooking the structured geometric information embedded in network parameters. In this work, we introduced a dual-manifold perspective that jointly characterizes the Kernel Manifold induced by network parameters and the Data Manifold formed by feature representations. Specifically, we proposed Kernel-Guided Feature Transform (KGFT), which explicitly transfers kernel geometry to feature space through a kernel-induced guidance matrix. Unlike conventional attention mechanisms that primarily reweight feature responses according to their estimated importance, KGFT explicitly reshapes feature relationships by transferring parameter-induced geometric structure into the data manifold. By combining Exploit and Explore modes with adaptive guidance strength and depth-aware scheduling, KGFT enables stable geometric alignment in shallow layers and promotes feature diversity in deeper layers, providing a lightweight and general mechanism for parameter-guided representation learning.

Extensive experiments across CNN- and Transformer-based architectures, including ResNet, ViT, and LLaMA-7B, demonstrate that KGFT consistently improves representation quality across image classification and arithmetic reasoning tasks with negligible parameter overhead. Theoretical analysis further establishes the validity, stability, and controllability of the proposed geometric transformation. These results suggest that network parameters are not merely optimization variables, but also contain structured geometric priors that can be explicitly exploited to complement conventional feature-centric representation learning.

In future work, we will further explore the application of KGFT to a broader range of downstream tasks, such as object detection, semantic segmentation, and parameter-efficient fine-tuning, to investigate whether kernel-guided geometry can provide consistent benefits beyond classification. More broadly, the dual-manifold principle offers a promising direction for cross-modal manifold alignment. For example, the Gram geometry of a text encoder ($\mathbf{G}_{text}$) could guide the covariance geometry of image features ($\mathbf{K}_{img}$), or vice versa, encouraging the principal directions and local tangent structures of different modalities to become geometrically compatible. Such a formulation could move beyond purely contrastive learning by explicitly aligning higher-order manifold geometry, providing a new perspective for multimodal representation learning.

\section*{Declaration of competing interest}
The authors declare that they have no known competing financial interests or personal relationships that could have appeared to influence the work reported in this paper.

\section*{Declaration of generative AI and AI-assisted technologies in the writing process}
During the preparation of this work the authors used ChatGPT in order to polish the language. After using this tool/service, the authors reviewed and edited the content as needed and took full responsibility for the content of the publication.

\section*{Acknowledgments}
This work was supported by a grant from the National Key R\&D Program of China (No.2024YFA1012002, No.2018YFC2001203), the General Program of the National Natural Science Foundation of China (No. 62471214), the AI Computing Platform, School of Biomedical Engineering, Southern Medical University.

\bibliographystyle{unsrtnat}
\bibliography{references}  

\section{Appendix A. Theoretical Analysis}
\subsection{The Geometric Interpretation of Exploit and Explore Guidance Matrices}
\textit{Exploit Geometric Interpretation: Stretch the data manifold along the principal direction of the kernel manifold. Let the eigenvalue decomposition of $\mathbf{G}$ be $\mathbf{G}=\sum_{i=1}^C\lambda_i\mathbf{u}_i\mathbf{u}_i^\mathsf{T}$; then, the scaling factor of $\mathbf{M_{exploit}}$ along the direction $\mathbf{u}_i$ is $\lambda_{i}+\varepsilon$ — where the direction with the longest kernel manifold stretch (i.e., the direction for which  is largest) corresponds to the greatest stretching of the data manifold.}

\textit{Explore Geometric Interpretation:Pushing the data manifold away from the principal directions of the kernel manifold. Let $\mathbf{C}=\sum_{i=1}^{C}\nu_{i}\mathbf{v}_{i}\mathbf{v}_{i}^{\top}$ be the eigenvalue decomposition of $\mathbf{C}$; then, the scaling factor of $\mathbf{M_{explore}}$ along direction $\mathbf{v}_{i}$ is given by $s(1-\mathbf{v}_{i})+\varepsilon$ — where directions on the kernel manifold with stronger correlations (i.e., $\mathbf{v}_{i}$) have smaller scaling factors.}

\subsection{Positive Definiteness of the Kernel-Induced Guidance}
\textbf{Proposition 1 (Positive Definiteness).} Let $\mathbf{G} = \mathbf{W}\mathbf{W}^\top$ be the kernel Gram matrix and let $\varepsilon > 0$. Then both $\mathbf{G}+\varepsilon\mathbf{I}_{C}$ and $\lambda\cdot(\mathbf{I}_C-\mathbf{C})+\varepsilon\mathbf{I}_C$ where $\mathbf{C}=\mathbf{D}^{-1/2}\mathbf{GD}^{-1/2}$ are symmetric positive definite (SPD). 

\textit{Proof.} Since $\mathbf{G}=\mathbf{W}\mathbf{W}^\top$ for any nonzero vector ($\mathbf{v}$):
\begin{equation}
	\mathbf{v}^\top\mathbf{G}\mathbf{v}=\mathbf{v}^\top\mathbf{W}\mathbf{W}^\top\mathbf{v}=\|\mathbf{W}^\top\mathbf{v}\|_2^2
\end{equation}
Therefore, $\mathbf{G}>\mathbf{0}$. For Exploit mode:
\begin{equation}
	\mathbf{v}^\top\mathbf{M}_\mathrm{exploit}\mathbf{v}=\mathbf{v}^\top\mathbf{G}\mathbf{v}+\varepsilon\parallel\mathbf{v}\parallel_2^2
\end{equation}
Because $\varepsilon>0$ and $\mathbf{v}\neq\mathbf{0}$, so $\mathbf{v}^\top\mathbf{M}_\mathrm{exploit}\mathbf{v}>0$. For Explore mode, let $\mathbf{G}=\mathbf{U\Lambda U}^{\mathrm{T}}$, where $\Lambda=\mathrm{diag}(\lambda_{1},\ldots,\lambda_{\mathrm{C}}),0\leq\lambda_{i}\leq\lambda_{\mathrm{max}}(\mathbf{G})$. The normalized matrix $\mathbf{C}=\mathbf{D}^{-1/2}\mathbf{GD}^{-1/2}$ has eigenvalues:
\begin{equation}
	\gamma_i=\frac{\lambda_i}{\lambda_{\max}(\mathbf{G})+\varepsilon}
\end{equation}
Since $\varepsilon > 0$, we get $0\leq\lambda_{i}\leq1$. Therefore, $\mathbf{0}<\mathbf{C}<\mathbf{I}$, and hence $(\mathbf{I}-\mathbf{C})>\mathbf{0}$. For any nonzero vector ($\mathbf{v}$):
\begin{equation}
	\mathbf{v}^\top\mathbf{M}_\mathrm{explore}\mathbf{v}=\delta\cdot\mathbf{v}^\top(\mathbf{I}-\mathbf{C})\mathbf{v}+\varepsilon\parallel\mathbf{v}\parallel_2^2>0
\end{equation}

\textbf{Conclusion:} This result directly justifies the guidance matrices introduced in Section 3.2.1. The SPD property guarantees that the kernel-induced operator defines a valid positive-definite metric in the shared channel space. Therefore, KGFT does not apply an arbitrary or potentially indefinite feature transformation; instead, its geometric modulation is induced by a mathematically valid metric.

\subsection{Lipschitz Stability of Kernel-Guided Mapping}
The kernel guidance depends on the network parameters through:
\begin{equation}
	\mathbf{W}\to\mathbf{G}=\mathbf{W}\mathbf{W}^\top\to\mathbf{M}(\mathbf{G})\to\mathbf{F}(\mathbf{X},\mathbf{M})
\end{equation}
Because network parameters continuously change during optimization, it is desirable that small parameter changes do not result in abrupt changes in the induced geometric transformation.

\textbf{Proposition 2 (Lipschitz Stability).} Assume that the kernel matrix satisfies: $\|\mathbf{W}\|_{2}\leq B_{W}$. Then the kernel Gram mapping $\mathbf{W}\to\mathbf{G}=\mathbf{W}\mathbf{W}^\top$ is Lipschitz continuous. Specifically, given two kernel matrices $\mathbf{w}_{1}$ and $\mathbf{w}_{1}$, we get $\|\mathbf{G}_1-\mathbf{G}_2\|_2\leq2B_W\|\mathbf{W}_1-\mathbf{W}_2\|_2$. Under bounded kernel scale and $\varepsilon > 0$, both the Exploit and Explore guidance matrices are Lipschitz continuous with respect to $\mathbf{W}$. Consequently, the resulting KGFT transformation is continuous with respect to perturbations of the kernel parameters.

\textit{Proof.} Given two kernel matrices, we calculate the difference: $\mathbf{W}_1\mathbf{W}_1^\top-\mathbf{W}_2\mathbf{W}_2^\top$. Adding and subtracting ( $\mathbf{W}_1\mathbf{W}_2^\top$), we obtain: 
\begin{equation}
	\mathbf{W}_1(\mathbf{W}_1-\mathbf{W}_2)^\top+(\mathbf{W}_1-\mathbf{W}_2)\mathbf{W}_2^\top
\end{equation}
Taking the spectral norm gives:
\begin{equation}
	\|\mathbf{G}_1-\mathbf{G}_2\|_2\leq\|\mathbf{W}_1\|_2\|\mathbf{W}_1-\mathbf{W}_2\|_2+\|\mathbf{W}_1-\mathbf{W}_2\|_2\|\mathbf{W}_2\|_2
\end{equation}
Since $\|\mathbf{W}_1\|_2,\|\mathbf{W}_1\|_2\leq B_W$, we obtain:
\begin{equation}
	\|\mathbf{G}_1-\mathbf{G}_2\|_2\leq2B_W\|\mathbf{W}_1-\mathbf{W}_2\|_2
\end{equation}
Thus, the kernel geometry changes continuously with the network parameters. Since the subsequent projection and normalization operations are continuous under standard boundedness assumptions, the complete KGFT mapping is also continuous with respect to $\mathbf{W}$.

\subsection{Bounded and Identity-Preserving Geometric Modulation}
The learnable guidance strength introduced in Section 3.2.3 controls the contribution of the geometrically transformed branch:
\begin{equation}
	\mathbf{Y}=\mathbf{X}+\mathbf{s}\cdot\mathcal{F}_{KGFT}(\mathbf{X})
\end{equation}
The following proposition explains why this residual formulation provides a bounded and controllable geometric modification.

\textbf{Proposition 3 (Boundedness and Identity Preservation).} Assume that the kernel-induced metric satisfies: $\|\mathbf{M}\|_2\leq B_M$, and that the geometric transformation branch satisfies: $\|\mathbf{F}(\mathbf{X},\mathbf{M})\|_F\leq B_F\|\mathbf{X}\|_F$, for some finite constant ($\leq B_F$). Then the KGFT output satisfies: $\|\mathbf{Y}\|_F\leq(1+\mathbf{s}B_F)\|\mathbf{X}\|_F$. Moreover, because $0<\mathbf{s}<1$, the geometric branch cannot dominate the residual pathway solely through an unconstrained guidance coefficient. In the limiting case $s\to0$, KGFT converges to the identity mapping: $\mathbf{Y}\to\mathbf{X}$.

\textit{Proof.} Let $\mathbf{Y}=\mathbf{X}+\mathbf{s}\cdot\mathcal{F}_{KGFT}(\mathbf{X})$, the triangle inequality gives:
\begin{equation}
	\|\mathbf{Y}\|_F\leq\|\mathbf{X}\|_F+\mathbf{s}\cdot\mathcal{F}_{KGFT}(\mathbf{X})
\end{equation}
Using the assumed bound: $\|\mathbf{F}(\mathbf{X},\mathbf{M})\|_F\leq B_F\|\mathbf{X}\|_F$, we get:
\begin{equation}
	\|\mathbf{Y}\|_F\leq\|\mathbf{X}\|_F+\mathbf{s}\cdot B_F\|\mathbf{X}\|_F
\end{equation}
and therefore obtain: 
\begin{equation}
	\|\mathbf{Y}\|_F\leq(1+\mathbf{s}\cdot B_F)\|\mathbf{X}\|_F
\end{equation}
Since $s=\sigma(\alpha)$, we have $0<\mathbf{s}<1$. Thus, the contribution of the geometric branch is explicitly bounded by its transformation magnitude rather than being controlled by an unconstrained scalar. Finally, when $\mathbf{s}>0,\mathbf{X}+\mathbf{s}\cdot\mathcal{F}_{KGFT}(\mathbf{X})\to\mathbf{X}$. Therefore, KGFT contains the identity mapping as a limiting case.

\textbf{Conclusion:} The role of $\mathbf{s}$ is not simply to introduce another learnable parameter; it provides an explicit mechanism for controlling the strength of geometric modulation. The identity-preserving property is particularly important in deep networks because it allows the optimization process to suppress the geometric branch when the injected kernel structure is not beneficial for a particular layer. Consequently, KGFT acts as an adaptive residual geometric modulation rather than a mandatory replacement of the original feature representation.

\section{Appendix B. Dual-manifold scheduling configuration}
Table \ref{table5} summarizes the detailed KGFT scheduling configurations across different architectures. For ResNet variants, KGFT is sparsely inserted into selected blocks across different stages, with Exploit generally used in shallow and intermediate stages and Explore introduced in deeper stages. For Transformer architectures, KGFT is similarly inserted into selected layers following the depth-aware scheduling principle. All KGFT modules use a learnable guidance strength. The specific insertion positions, operating modes, and guidance settings are provided in Table \ref{table5}.
\begin{table}[t]
	\centering
	\small
	\renewcommand{\arraystretch}{1.2}
	\caption{The detailed configurations of KGFT scheduling strategy on ResNet and Transformer architectures.}
	\label{table5}
	\begin{tabular}{l c c c c c}
		\toprule
		\textbf{Architecture} & \textbf{Stages} & \textbf{Blocks} & \textbf{KGFT Mode} & \textbf{Position} & \textbf{Strength} \\
		\midrule
		\multirow{4}{*}{ResNet-18} & Stage1 & 2 & Exploit & Block 2 & Learnable \\
		& Stage2 & 2 & Exploit & Block 2 & Learnable \\
		& Stage3 & 2 & Exploit & Block 2 & Learnable \\
		& Stage4 & 2 & Explore & Block 2 & Learnable \\
		\midrule
		\multirow{3}{*}{ResNet-20} & Stage1 & 3 & Exploit & Block 3 & Learnable \\
		& Stage2 & 3 & Exploit & Block 3 & Learnable \\
		& Stage3 & 3 & Explore & Block 3 & Learnable \\
		\midrule
		\multirow{3}{*}{ResNet-32} & Stage1 & 5 & Exploit & Block 3 \& 5 & Learnable \\
		& Stage2 & 5 & Exploit & Block 3 \& 5 & Learnable \\
		& Stage3 & 5 & Explore & Block 3 \& 5 & Learnable \\
		\midrule
		\multirow{4}{*}{ResNet-34} & Stage1 & 3 & Exploit & Block 3 & Learnable \\
		& Stage2 & 4 & Exploit & Block 2 \& 4 & Learnable \\
		& Stage3 & 6 & Exploit & Block 3 \& 6 & Learnable \\
		& Stage4 & 3 & Explore & Block 3 & Learnable \\
		\midrule
		\multirow{4}{*}{ResNet-50} & Stage1 & 3 & Exploit & Block 3 & Learnable \\
		& Stage2 & 4 & Exploit & Block 2 \& 4 & Learnable \\
		& Stage3 & 6 & Exploit & Block 3 \& 6 & Learnable \\
		& Stage4 & 3 & Explore & Block 3 & Learnable \\
		\midrule
		\multirow{3}{*}{ViT-tiny}  & -- & L12 & Exploit & L4 & Learnable \\
		& -- & L12 & Exploit & L8 & Learnable \\
		& -- & L12 & Explore & L12 & Learnable \\
		\midrule
		\multirow{3}{*}{LLaMA-7B}  & -- & L32 & Exploit & L9 & Learnable \\
		& -- & L32 & Exploit & L21 & Learnable \\
		& -- & L32 & Exploit & L32 & Learnable \\
		\bottomrule
	\end{tabular}
\end{table}

\begin{figure*}[t]
	\centering
	\includegraphics[width=0.9\columnwidth]{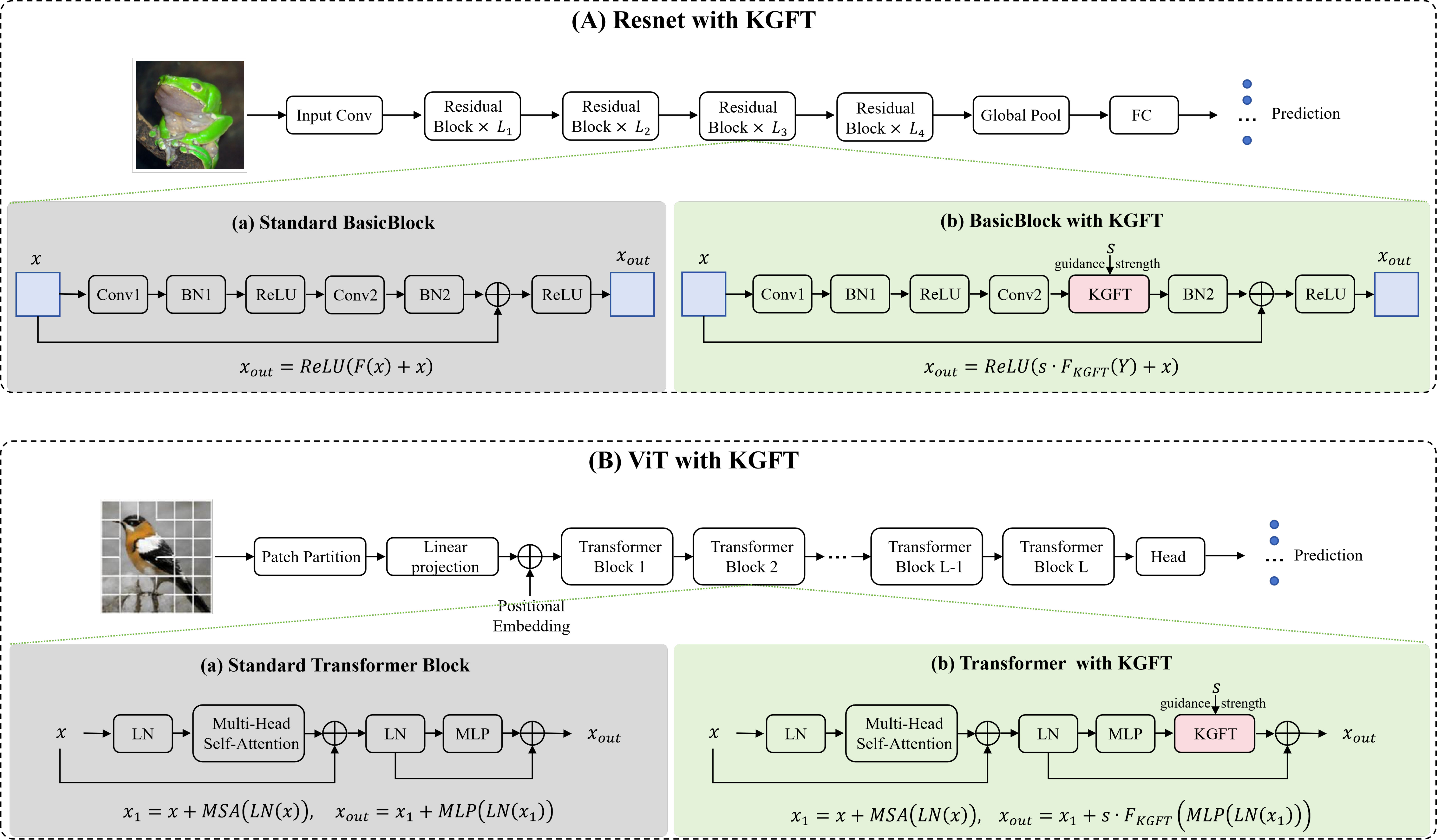}
	\caption{Overview of KGFT applied to ResNet and ViT architectures. The upper panels illustrate the overall architectures of ResNet and ViT equipped with KGFT, where KGFT is selectively inserted into representative network blocks according to the proposed depth-aware dual-manifold scheduling strategy.}
	\label{fig2}
\end{figure*}

Figure \ref{fig2} provides a visual illustration of KGFT integration into ResNet and ViT. In ResNet, KGFT is inserted within selected residual blocks and fused with the residual pathway, while in ViT, it is applied after the MLP sublayer and incorporated through the residual connection. These implementations preserve the original backbone architecture while enabling KGFT to function as a lightweight plug-and-play module.

\end{document}